\documentclass[letterpaper, 10 pt, conference]{ieeeconf}  

\IEEEoverridecommandlockouts                              

\let\labelindent\relax 
\usepackage{enumitem}
\usepackage{xcolor}
\usepackage{diagbox}
\usepackage{multirow}
 
\usepackage{amsmath}
\usepackage{amsthm,amssymb}
\usepackage{graphicx}
\newtheorem{theorem}{Theorem}

\newtheorem{lemma}{Lemma}
\newtheorem{remark}{Remark}
\newtheorem{defi}{Definition}

\newcommand{\bR}{\mathbb{R}}

\newcommand{\cA}{\mathcal{A}}

\newcommand{\cP}{\mathcal{P}}

\newcommand{\cS}{\mathcal{S}}

\newcommand{\rx}{r_{\max}}

\newcommand{\na}{\textnormal{(a)}}
\newcommand{\nb}{\textnormal{(b)}}
\newcommand{\nc}{\textnormal{(c)}}

\newcommand{\esr}[1]{{\color{orange}ESR: #1}}

\usepackage{comment}
\usepackage{algorithm}
\usepackage{algorithmic}

\title{\LARGE \bf
Teaching Precommitted Agents: Model-Free Policy Evaluation and Control in Quasi-Hyperbolic Discounted MDPs
}

\author{ S.R. Eshwar
\thanks{S.R. Eshwar is with Department of Computer Science and Automation, Indian Institute of Science, Bengaluru, India.  He was supported by the Prime Minister’s Research Fellowship (PMRF). This work was partially supported by the Walmart Centre for Tech Excellence at IISc. This work was partially supported by a grant from the National Payments Corporation of India (NPCI) to IISc.
{\tt\small eshwarsr@iisc.ac.in}}%
}

\begin{document}

\maketitle
\thispagestyle{empty}
\pagestyle{empty}

\begin{abstract}
Time-inconsistent preferences, where agents favor smaller-sooner over larger-later rewards, are a key feature of human and animal decision-making. Quasi-Hyperbolic (QH) discounting provides a simple yet powerful model for this behavior, but its integration into the reinforcement learning (RL) framework has been limited. This paper addresses key theoretical and algorithmic gaps for precommitted agents with QH preferences. We make two primary contributions: (i) we formally characterize the structure of the optimal policy, proving for the first time that it reduces to a simple one-step non-stationary form; and (ii) we design the first practical, model-free algorithms for both policy evaluation and Q-learning in this setting, both with provable convergence guarantees. Our results provide foundational insights for incorporating QH preferences in RL.
\end{abstract}

\section{Introduction}

Reinforcement learning (RL) \cite{sutton2018reinforcement} provides a powerful framework for sequential decision-making, where an agent interacts with an environment to maximize long-term cumulative rewards. A foundational assumption in most RL algorithms is that agents discount future rewards using an exponential discount factor, which implies time-consistent preferences: an agent's relative preference between two future rewards remains unchanged over time. However, this assumption does not hold in many real-world scenarios involving humans, whose preferences are known to be time-inconsistent. \cite{dhami2016foundations}.

Quasi-Hyperbolic (QH) discounting \cite{phelps1968second, Laib}, introduced in behavioral economics, offers a more realistic model of decision-making. Under QH discounting, agents apply a present-bias parameter \( \sigma \in [0,1] \) in addition to the standard exponential discount factor \( \gamma \in [0,1) \). Immediate rewards are weighted more heavily than future rewards, capturing the observed preference for smaller-sooner outcomes over larger-later ones. Formally, the cumulative discounted return for a sequence of rewards \( (r_0, r_1, r_2, \dots) \) is given by:
\begin{equation*}
    G = r_0 + \sigma \sum_{t=1}^{\infty} \gamma^t r_t.
\end{equation*}

While QH discounting has been used to explain human behavior in intertemporal choice tasks, its incorporation into the RL framework poses unique challenges. Unlike exponential discounting, QH discounting leads to time-inconsistent preferences: the optimal policy from the perspective of the present self may differ from what a future self would prefer. This gives rise to a spectrum of agent behaviors \cite{naive_vs_sophesticate,strotz1955myopia}.

\begin{itemize}
    \item \textbf{Naive agents} fail to recognize their future time inconsistency and continually replan.
    \item \textbf{Sophisticated agents} \cite{mdp_with_qh, rl_with_qh_mpe} anticipate their future selves' behavior and seek subgame perfect equilibria.
    \item \textbf{Precommitted agents} understand their time inconsistency but fix their policies in advance to optimize cumulative QH-discounted return.
\end{itemize}

This paper centers on the precommitted agent \cite{strotz1955myopia}, a model capturing decision-makers who can identify and commit to a long-term optimal policy from the outset. This widely-studied paradigm is well-motivated by real-world scenarios where individuals utilize `commitment devices'—such as setting deadlines or creating savings plans—to adhere to a long-term strategy. Adopting this framework provides a clear and tractable optimization problem, which allows us to formally characterize the structure of the globally optimal policy under QH discounting. This model is therefore fundamental for understanding optimal planning with time-inconsistent preferences, serving as an essential benchmark against which more complex, strategic behaviors can be compared.

Our work on quasi-hyperbolic (QH) discounting is distinct from other prominent frameworks for modeling non-standard preferences in MDPs. One such framework is risk-sensitive RL, which captures human-like attitudes toward outcome uncertainty \cite{prashanth2022risk}. Operationally, the goal is to find a policy that either optimizes a risk measure as an objective or optimizes the usual risk-neutral objective while satisfying a risk constraint. This focus on managing risk addresses a fundamentally different problem than our work, which seeks to model human-like temporal preferences.

Another related framework is reinforcement learning with prospect theory \cite{borkar2021prospect}. Prospect theory describes how individuals assess losses and gains asymmetrically, driven by loss aversion—the principle that the pain of a loss is felt more strongly than the pleasure of an equivalent gain. This asymmetry explains why agents may be risk-averse when facing potential gains but risk-seeking to avoid losses. While this also leads to complex objectives, it models behavioral attitudes toward risk and value, which is distinct from the human-like preferences over the timing of rewards captured by QH discounting.

\noindent\textbf{Contributions:} Our primary contributions are threefold.

\begin{enumerate}
    \item We characterize the structure of optimal policies under QH discounting for precommitted agents. To the best of our knowledge, we are the first to show that the optimal policy is a \emph{one-step non-stationary policy}. That is, the optimal policy can be decomposed into a pair \( (\mu^*, \bar{\pi}^*) \), where \( \mu^* \) is applied at the first step, and \( {\pi}^* \) is a stationary policy used thereafter.

    \item We develop a provably convergent, model-free algorithm for the off-policy evaluation of any given one-step non-stationary policy.
    \item We design QH Q-Learning, a novel model-free algorithm that learns the optimal policy by concurrently estimating the optimal QH and exponential action-value functions.
\end{enumerate}



\noindent\textbf{Organization:} The remainder of this paper is organized as follows. Section~\ref{sec:setup_prelim} formalizes the MDP with QH discounting and presents our theoretical result on the one-step non-stationary representation of any non-stationary policy. Section~\ref{sec:policy_evaluation} presents the policy evaluation algorithm for any given one-step non-stationary policy. Section~\ref{sec:optimal_control} details the structure of the optimal policy and our corresponding QH Q-Learning algorithm. Finally, Section~\ref{sec:experiments} provides an empirical validation of our methods, followed by our conclusion.

\section{Setup and Preliminaries}
\label{sec:setup_prelim}

\subsection{Setup}
Let $\Delta(\mathcal{U})$ denote the set of probability distributions over a set $\mathcal{U}$. Our setup consists of a Markov Decision Process (MDP) $ M \equiv (\mathcal{S}, \mathcal{A}, \mathcal{P}, r, \sigma, \gamma), $ where:
$\mathcal{S}$ is a finite state space, $\mathcal{A}$ is a finite action space, $\mathcal{P}: \mathcal{S} \times \mathcal{A} \to \Delta(\mathcal{S})$ is the transition probability kernel, $r: \mathcal{S} \times \mathcal{A} \to \mathbb{R}$ is the reward function, $\sigma, \gamma \in [0, 1)$ are the parameters of quasi-hyperbolic (QH) discounting.

The QH state-value and action-value functions for a policy $\nu$, denoted $V^{\sigma,\gamma}_\nu$ and $Q^{\sigma,\gamma}_\nu$ respectively, are defined as follows:
\begin{equation*}
\begin{split}
    V^{\sigma,\gamma}_{\nu}(s) &= \mathbb{E} \left[ \sum_{t=0}^{\infty} d(t) r(s_t, a_t) \mid s_0 = s \right],\\
    Q^{\sigma,\gamma}_{\nu}(s,a) &= \mathbb{E} \left[ \sum_{t=0}^{\infty} d(t) r(s_t, a_t) \mid s_0 = s,a_0=a \right]
\end{split}
\end{equation*}
where the QH discount factor \( d(t) \) is given by  
\begin{equation*}
    d(t) =
    \begin{cases} 
        1, & t = 0, \\ 
        \sigma \gamma^t, & t \geq 1.
    \end{cases}
\end{equation*}

For the special case where \( \sigma = 1 \), the QH formulation reduces to standard exponential discounting, with corresponding value functions denoted by \( V^{\gamma}_\nu \) and \( Q^{\gamma}_\nu \).

\subsection{One-Step Non-Stationary Representation}
\label{sec:one_step_non_stat}
In this section, we introduce a key structural property of policies under quasi-hyperbolic discounting. A well-known fact in exponential discounting is that the value function of any non-stationary policy is equivalent to the value function of a corresponding stationary policy. In a similar spirit, we establish a new result for quasi-hyperbolic discounting: any non-stationary policy can be equivalently represented as a \emph{one-step non-stationary policy}, where a policy $\mu$ is followed in the first step, after which a stationary policy $\pi$ is followed indefinitely, represented as $(\mu,\bar{\pi}) = (\mu, \pi, \pi, ...)$.

\begin{theorem}
For a finite-state, finite-action MDP, the value function under quasi-hyperbolic discounting of any non-stationary policy $\nu = (\nu_0, \nu_1, ...)$ is equal to the value function corresponding to a one-step non-stationary policy $(\mu, \bar{\pi})$ for some choice of policies $\mu$ and $\pi$.
\end{theorem}

\begin{proof}
From the definition of QH-discounted value functions,
\begin{equation*}
    V^{\sigma,\gamma}_{\nu}(s) = \mathbb{E} \left[ r(s_0, a_0) + \sum_{t=1}^{\infty} \sigma \gamma^t r(s_t, a_t) \mid s_0 = s \right].
\end{equation*}
Rewriting the expectation by conditioning on the first action \( a_0 \sim \nu_0(\cdot | s) \), we get:  
\begin{equation*}
    V^{\sigma,\gamma}_{\nu}(s) = \mathbb{E} \left[ r(s, a_0) + \mathbb{E} \left[ \sum_{t=1}^{\infty} \sigma \gamma^t r(s_t, a_t) \mid s_0, a_0 \right] \right].
\end{equation*}
Further conditioning on the next state \( s' \sim \cP(\cdot | s, a_0) \), we obtain:  
\begin{equation*}
    V^{\sigma,\gamma}_{\nu}(s) = \mathbb{E}\left[ r(s, a_0) + \sigma \gamma \mathbb{E} \left[ \sum_{t=1}^{\infty}  \gamma^{t-1} r(s_t, a_t) \mid s_1 = s' \right] \right]
\end{equation*}
Using the fact that the remaining discounted sum follows an exponentially discounted value function,  
\begin{equation*}
    \mathbb{E} \left[ \sum_{t=1}^{\infty}  \gamma^{t-1} r(s_t, a_t) \mid s_1 = s' \right] = V^{\gamma}_{(\nu_1, \nu_2, ...)}(s'),
\end{equation*}
we conclude:  
\begin{equation}
\label{eq:V_qn_in_terms_of_V_exp}
    V^{\sigma,\gamma}_{\nu}(s) = \mathbb{E}_{\substack{a_0 \sim \nu_0(\cdot | s)\\ s' \sim \cP(\cdot | s, a_0)}} \left[ r(s, a_0) + \sigma \gamma V^{\gamma}_{(\nu_1, \nu_2, ...)}(s') \right].
\end{equation}

It is well-established that for any non-stationary policy under exponential discounting, an equivalent stationary policy can be constructed that produces an identical value function \cite[Prop. 1.1.1]{bertsekas_dp_oc}. Accordingly, let $\bar{\pi}$ be a stationary policy such that:
\begin{equation*}
    V^{\gamma}_{(\nu_1, \nu_2, ...)}(s) = V^{\gamma}_{\bar{\pi}}(s), \quad \forall s \in \mathcal{S}.
\end{equation*}

Substituting this into the previous equation, we obtain:
\begin{equation*}
    V^{\sigma,\gamma}_{\nu}(s) = \sum_a \nu_0(a | s) \Big[r(s, a) + \sigma \gamma \sum_{s'} \cP(s' | s, a) V^{\gamma}_{\bar{\pi}}(s')\Big].
\end{equation*}

Defining $\mu = \nu_0$ as the first-step policy, we conclude that
\begin{equation*}
    V^{\sigma,\gamma}_{\nu}(s) = V^{\sigma,\gamma}_{\mu, \bar{\pi}}(s),
\end{equation*}
thus proving the theorem.
\end{proof}

\section{Policy Evaluation}
\label{sec:policy_evaluation}

In this section, we develop a model-free algorithm to evaluate the value function of a one-step non-stationary policy, $(\mu, \bar{\pi})$, under quasi-hyperbolic (QH) discounting (see Algorithm \ref{alg:policy_evaluation}). As established in Section \ref{sec:one_step_non_stat}, this policy class is sufficient for our analysis. A key challenge in QH discounting is that the value function $V^{\sigma,\gamma}_{\mu,\bar{\pi}}$ does not satisfy a direct Bellman-style recursion. We resolve this issue by first deriving a relationship that expresses $V^{\sigma,\gamma}_{\mu,\bar{\pi}}(s)$ in terms of the value function of stationary component, $V^{\sigma,\gamma}_{\bar{\pi}}(s)$, which can itself be defined recursively.

The value function for the policy $(\mu, \bar{\pi})$ is defined as
$$V_{\mu,\bar{\pi}}^{\sigma, \gamma}(s) = \mathbb{E}\left[r(s_0,a_0) + \sigma \sum_{t=1}^{\infty} \gamma^t r(s_t, a_t) \mid s_0 = s\right],$$
where the expectation is taken over trajectories generated by acting with $\mu$ at the first step and $\pi$ thereafter. Through algebraic manipulation, we can decompose this expectation to derive a one-step lookahead expression for $V_{\mu,\bar{\pi}}^{\sigma, \gamma}(s)$.
\begin{align*}
    &V_{\mu,\bar{\pi}}^{\sigma, \gamma}(s)\\
    &= \mathbb{E}\big[r(s_0,a_0) + \sigma \sum_{t=1}^{\infty} \gamma^t r(s_t, a_t) \mid s_0 = s\big] \\
    &= \mathbb{E}\Bigg[r(s_0, a_0) + \sigma \gamma r(s_1, a_1)  - \gamma r(s_1,a_1) \\
    & \quad \quad + \gamma \left(r(s_1,a_1) + \sigma  \sum_{t=2}^{\infty} \gamma^{t-1} r(s_t, a_t) \right) \mid s_0 = s\Bigg] \\
    &= \sum_a \mu(a|s) \Bigg[r(s,a) +  \sum_{s'} \cP(s'|s,a) \Big[-(1-\sigma)\gamma \\
    & \quad \sum_{a'} \pi(a'|s') r(s',a') + \gamma V^{\sigma,\gamma}_{\bar{\pi}}(s')\Big] \Bigg]
\end{align*}

By specializing this result to the case where $\mu = \pi$, we obtain a self-consistent Bellman equation for the stationary policy value function $V^{\sigma,\gamma}_{\bar{\pi}}(s)$.
\begin{align*}
    &V^{\sigma,\gamma}_{\bar{\pi}}(s) = \sum_a \pi(a|s) \Bigg[r(s,a) + \sum_{s'} \cP(s'|s,a) \Big[-(1-\sigma) \\
    & \quad \gamma\sum_{a'} \pi(a'|s') r(s',a') + \gamma V^{\sigma,\gamma}_{\bar{\pi}}(s')\Big] \Bigg]
\end{align*}

Since the evaluation of both \( V^{\sigma,\gamma}_{\mu,\bar{\pi}} \) and \( V^{\sigma,\gamma}_{\bar{\pi}} \) requires sampling actions according to the policies \(\mu\) and \(\pi\) respectively, we adopt an off-policy approach. We introduce a behavior policy $\nu$ to generate data and use importance sampling to correct for the mismatch between this policy and the target policies. This requires the standard coverage assumption: for any state $s$, $\nu(a|s) > 0$ whenever $\mu(a|s) > 0$ or $\pi(a|s) > 0$. To this end, our algorithm maintains two iterates to estimate the value functions (i) \( W_n(s) \), which estimates \( V^{\sigma,\gamma}_{\bar{\pi}}(s) \), and  
(ii) \( V_n(s) \), which estimates \( V^{\sigma,\gamma}_{\mu,\bar{\pi}}(s) \). See Algorithm \ref{alg:policy_evaluation} for more details.

\begin{algorithm}[tb]
   \caption{Model-Free Policy Evaluation for QH Discounting
   }
   \label{alg:policy_evaluation}
\begin{algorithmic}[1]
   \STATE {\bfseries Input:} Stepsizes $(\alpha_n)$; discount factors $\sigma, \gamma$; sampling policy $\nu$; evaluation policies $\mu, \pi$
   \STATE {\bfseries Initialize:} $V_0, W_0 \in \mathbb{R}^{|\mathcal{S}|}$
   \FOR{$n = 0, 1, 2, \ldots$}
       \STATE Set $D_n^{(1)} = D_n^{(2)} = 0 \in \mathbb{R}^{|\cS|\times|\cS|}, W_n^{\text{target}} = 0 \in \mathbb{R}^{|\cS|}$
       \FOR{$s \in \mathcal{S}$}
           \STATE Sample $a \sim \nu(\cdot|s)$
           \STATE Observe reward $r(s,a)$ and next state $s' \sim \cP(\cdot|s,a)$
           \STATE Sample $a' \sim \pi(\cdot|s')$
           \STATE Observe reward $r(s', a')$
           \STATE $W_n^{\text{target}}(s) \leftarrow r(s,a) - (1 - \sigma)\gamma r(s', a') + \gamma W_n(s')$
           \STATE $D_n^{(1)}(s,s) \leftarrow \frac{\pi(a|s)}{\nu(a|s)},  D_n^{(2)}(s,s) \leftarrow \frac{\mu(a|s)}{\nu(a|s)}$
           \ENDFOR
       \STATE Update $W$: 
       \[
       W_{n+1} \leftarrow W_n + \alpha_n \left[D_n^{(1)} W_n^{\text{target}} - W_n\right]
       \]
       \STATE Update $V$: 
       \[
       V_{n+1} \leftarrow V_n + \alpha_n \left[D_n^{(2)} W_n^{\text{target}} - V_n\right]
       \]
   
   \ENDFOR
\end{algorithmic}
\end{algorithm}

We now state our assumptions, followed by the main result of this section. 
{
\renewcommand{\theenumi}{$\cA_\arabic{enumi}$}
\begin{enumerate}[leftmargin=*]
\item \label{a:stepsize} \textbf{Stepsizes: }$(\alpha_n)_{n \geq 0}$ are positive scalars satisfying 
        (i) $\alpha_0 \leq 1;$ 
        (ii) $\sum_{n=0}^{\infty} \alpha_n=\infty,$ but $\sum_{n=0}^{\infty} \alpha_n^2 <\infty.$
\item \label{a:reward} \textbf{Bounded reward:} There exists $\rx > 0$ such that $|r(s,a)| \leq \rx$ for all $s \in \cS$ and $a \in \cA.$ 
\item \label{a:bdd_is_ratio} \textbf{Bounded sampling ratios:} the importance sampling ratios are uniformly bounded, such that
$\rho^{(1)}_{\max} := \max_{s,a} \frac{\pi(a|s)}{\nu(a|s)} < \infty$ and $\rho^{(2)}_{\max} := \max_{s,a} \frac{\mu(a|s)}{\nu(a|s)} < \infty.$

\end{enumerate}
}

\begin{theorem}[Convergence of Policy Evaluation]
\label{th:policy_evaluation}
Suppose \ref{a:stepsize}, \ref{a:reward} and \ref{a:bdd_is_ratio} are true.
Then the iterates \( (W_n, V_n) \) obtained from Algorithm~\ref{alg:policy_evaluation} converge almost surely to the true value functions \( (V^{\sigma,\gamma}_{\bar{\pi}}, V^{\sigma,\gamma}_{\mu,\bar{\pi}}) \).
\end{theorem}
\begin{proof}[Proof Sketch]
Our proof relies on the ODE method from stochastic approximation \cite{borkar_sa_book}. The analysis first confirms that the iterates are bounded. We then show that these iterates asymptotically track a limiting ODE whose unique, globally asymptotically stable equilibrium is precisely the true value function. This stability property, combined with the Lipschitz continuity of the underlying dynamics and standard noise assumptions, allows us to invoke established convergence theorems to conclude the proof.
\end{proof}




\section{Optimal Control}
\label{sec:optimal_control}

In this section, we characterize the optimal policy under quasi-hyperbolic (QH) discounting and provide the foundation for a model-free algorithm to compute it.

\subsection{Structure of the Optimal Policy}

We begin by defining the optimal value function $V^*$ as the maximum achievable value from any state, and then derive its structure.

\begin{defi}
The optimal value function in QH discounting, $V^*$, satisfies:
\begin{equation*}
    V^*(s) = \max_{\mu,\pi} V^{\sigma,\gamma}_{\mu,\bar{\pi}}(s), \quad \forall s \in \mathcal{S}.
\end{equation*}
An optimal policy $(\mu^*, \bar{\pi}^*)$ is any policy that achieves this maximum.
\end{defi}

To find the structure of this optimal policy, we can expand the definition of $V^*$ using the relationship between the QH value function and the standard exponential value function from \eqref{eq:V_qn_in_terms_of_V_exp}. This allows us to derive a Bellman-style optimality equation:
\begin{equation*}
    \begin{split}
        &V^*(s) \\
        &= \max_{\mu,\pi} \sum_a \mu(a|s) \left[ r(s,a) + \sigma\gamma \sum_{s'} \cP(s'|s,a) V^{\gamma}_{\bar{\pi}}(s') \right] \\
        &= \max_{\mu} \sum_a \mu(a|s) \left[ r(s,a) + \sigma\gamma \sum_{s'} \cP(s'|s,a)  \max_{\pi} V^{\gamma}_{\bar{\pi}}(s') \right]
    \end{split}
\end{equation*}
The inner term, $\max_{\pi} V^{\gamma}_{\bar{\pi}}(s')$, is precisely the optimal value function for a standard MDP with exponential discounting, which we denote as $V_*^{\gamma}(s')$. The policy that achieves this is the standard optimal policy, $\pi_*^{\gamma}$.

This decomposition reveals two key insights about the structure of the optimal QH policy $(\mu^*, \bar{\pi}^*)$:
\begin{enumerate}
    \item \textbf{Optimal Stationary Component:} The optimal policy for all future steps, $\bar{\pi}^*$, must be the policy that is optimal for the standard exponential discounting problem. Thus,
    $$\bar{\pi}^* = \pi_*^{\gamma}.$$
    \item \textbf{Optimal Initial Policy:} The optimal policy for the first step, $\mu^*$, must greedily select the action that maximizes the immediate reward plus the discounted value of acting optimally thereafter. This implies that $\mu^*$ is the greedy policy with respect to the following action-value function
    \begin{equation}
    \label{eq:Q_hybrid_def}
        Q^{\sigma,\gamma}_*(s,a) := r(s,a) + \sigma\gamma \sum_{s'} \cP(s'|s,a) V_*^{\gamma}(s').
    \end{equation}
\end{enumerate}
\subsection{Existence of a Deterministic Optimal Policy}

With the structure of the optimal policy established, we can now state the following theorem.

\begin{theorem}
An optimal policy $(\mu^*, \bar{\pi}^*)$ for the quasi-hyperbolic discounting problem exists and can be chosen to be deterministic.
\end{theorem}
\begin{proof}
The structure of the optimal policy $(\mu^*, \bar{\pi}^*)$ was derived above. We prove that each component can be deterministic.
\begin{enumerate}
    \item From standard MDP theory, an optimal policy under exponential discounting, $\pi_*^{\gamma}$, is known to exist and be deterministic. Since the optimal stationary component is $\bar{\pi}^* = \pi_*^{\gamma}$, it follows that a deterministic $\bar{\pi}^*$ can be chosen.
    \item The optimal initial policy $\mu^*$ is the greedy policy with respect to $Q^{\sigma,\gamma}_*(s,a)$, meaning $\mu^*(a|s) = 1$ for $a \in \arg\max_{a'} Q^{\sigma,\gamma}_*(s,a')$ and $0$ otherwise. The $\arg\max$ operator produces a deterministic policy (with ties broken arbitrarily). Therefore, $\mu^*$ is also deterministic.
\end{enumerate}
This establishes the existence of a deterministic optimal policy $(\mu^*, \bar{\pi}^*)$.
\end{proof}

\begin{remark}
This result is significant as it guarantees that the search for an optimal policy can be restricted from the vast space of all non-stationary stochastic policies to the much smaller space of one-step non-stationary deterministic policies.
\end{remark}

\subsection{Model-Based Solution via Value Iteration}
\label{sec:val_iter}

The structure of the optimal policy $(\mu^*, \bar{\pi}^*)$ derived in the previous section provides a clear blueprint for a model-based algorithm. This approach, analogous to standard value iteration, consists of two stages.

First, since the optimal stationary component $\bar{\pi}^*$ is equivalent to the optimal policy under exponential discounting ($\pi_*^\gamma$), we can compute its corresponding value function, $V_*^{\gamma}$, using the standard Bellman optimality operator until convergence:
\begin{equation*}
    V_{k+1}(s) \leftarrow \max_{a \in \mathcal{A}} \left[ r(s,a) + \gamma \sum_{s' \in \mathcal{S}} \cP(s'|s,a) V_k(s') \right].
\end{equation*}
Once $V_*^{\gamma}$ is obtained, the second stage is to compute the optimal initial policy $\mu^*$. This is done by performing a one-step greedy policy extraction with respect to the action-value function defined in \eqref{eq:Q_hybrid_def}:
\begin{equation*}
    \mu^*(s) = \arg\max_{a \in \mathcal{A}} Q^{\sigma,\gamma}_*(s,a).
\end{equation*}

While this two-stage procedure is straightforward when the model dynamics $\cP(s'|s,a)$ and reward function $r(s,a)$ are known, our primary interest lies in the more practical model-free setting. In a model-free context, we cannot perform the one-step lookahead required by the Bellman operator or the computation of $Q^{\sigma,\gamma}_*$. This limitation motivates a shift from state-value functions to action-value functions, which enable policy optimization without a model. Accordingly, we develop a model-free Q-learning algorithm for quasi-hyperbolic discounting in the next section.

\subsection{Model-Free Algorithm}
\label{sec:model_free_alg}

\begin{algorithm}[tb]
   \caption{QH Q-learning (Model-Free)}
   \label{alg:qh_Q_learning}
\begin{algorithmic}[1]
   \STATE {\bfseries Input:} Stepsizes $(\alpha_n)$, discount factors $\sigma$ and $\gamma$
   \STATE {\bfseries Initialize:} $Z_0, Q_0 \in \mathbb{R}^{|\mathcal{S}||\mathcal{A}|}$
   \FOR{$n = 0, 1, 2, \ldots$}
       \STATE Set $Z'_n = 0 \in \mathbb{R}^{|\cS| |\cA|}$
       \FOR{$(s,a) \in \mathcal{S} \times \mathcal{A}$}
           \STATE Sample $s' \sim \cP(\cdot|s,a)$
           \STATE Observe reward $r(s,a)$
           \STATE $Z'_n(s,a) \leftarrow \max_{b} Z_n(s',b)$
        \ENDFOR
       \STATE $Z_{n+1} \leftarrow Z_n + \alpha_n \big[ r + \gamma Z'_n - Z_n \big]$
       \STATE $Q_{n+1} \leftarrow Q_n + \alpha_n \big[ (1-\sigma)r + \sigma Z_n - Q_n \big]$
   \ENDFOR
\end{algorithmic}
\end{algorithm}

In this section, we extend the preceding model-based analysis to develop a practical, model-free algorithm for computing the optimal policy. We present Algorithm \ref{alg:qh_Q_learning}, which learns the necessary action-value functions without requiring knowledge of the system dynamics.

\subsubsection{Relationship between QH and Exponential Q-Functions}

We begin by formalizing the relationship between the quasi-hyperbolic and exponential Q-functions. We follow the standard convention where the policy subscript in $Q(s,a)$ denotes the policy executed after taking action $a$. For a one-step non-stationary policy $(\mu, \bar{\pi})$, once the initial action $a$ is taken, the subsequent policy is always the stationary component $\bar{\pi}$. Thus, the action-value function is denoted as $Q^{\sigma,\gamma}_{\bar{\pi}}(s,a)$, which can be decomposed by expanding its definition:
\begin{align*}
    &Q^{\sigma,\gamma}_{\bar{\pi}}(s,a)\\ &=\mathbb{E}\left[r(s_0, a_0) + \sigma \sum_{t=1}^{\infty} \gamma^t r(s_t, a_t) \mid s_0=s, a_0=a\right] \\
    &= \mathbb{E}\Bigg[ (1-\sigma)r(s_0, a_0) \\
    & \quad \quad + \sigma r(s_0, a_0) + \sigma \sum_{t=1}^{\infty} \gamma^t r(s_t, a_t) \mid s_0=s, a_0=a \Bigg] \\
    &= (1-\sigma)r(s,a) + \sigma \, \mathbb{E}\left[ \sum_{t=0}^{\infty} \gamma^t r(s_t, a_t) \mid s_0=s, a_0=a \right] \\
    &= (1-\sigma)r(s,a) + \sigma Q^{\gamma}_{\bar{\pi}}(s,a).
\end{align*}
This yields the key linear relationship that our algorithm will exploit:
\begin{equation}
\label{eq:q_sigma_q_gamma_relation}
    Q^{\sigma,\gamma}_{\bar{\pi}}(s,a) = (1-\sigma)r(s,a) + \sigma Q^{\gamma}_{\bar{\pi}}(s,a).
\end{equation}

\subsubsection{Optimality Equation for QH Q-Functions}

Building upon the linear relationship established in \eqref{eq:q_sigma_q_gamma_relation}, we can now derive the optimality equation for the quasi-hyperbolic setting. The optimal action-value function, $Q^{\sigma,\gamma}_*(s,a)$, is found by maximizing both sides of \eqref{eq:q_sigma_q_gamma_relation} over all stationary component policies $\pi$ and initial policies $\mu$:
\begin{equation*}
    \begin{split}
        Q^{\sigma,\gamma}_*(s,a) &:= \max_{\mu, \pi} Q^{\sigma,\gamma}_{\bar{\pi}}(s,a) \\
        &= \max_{\pi} \left[ (1-\sigma)r(s,a) + \sigma Q^{\gamma}_{\bar{\pi}}(s,a) \right] \\
        &= (1-\sigma)r(s,a) + \sigma \left( \max_{\pi} Q^{\gamma}_{\bar{\pi}}(s,a) \right).
    \end{split}
\end{equation*}
We identify the term $\max_{\pi} Q^{\gamma}_{\bar{\pi}}(s,a)$ as the optimal action-value function from standard exponential discounting, denoted $Q^{\gamma}_*(s,a)$. This yields the optimality equation for QH Q-values:
\begin{equation}
\label{eq:qh_q_bellman_optimality}
    Q^{\sigma,\gamma}_*(s,a) = (1-\sigma)r(s,a) + \sigma Q^{\gamma}_*(s,a).
\end{equation}
This decomposition reveals a crucial insight: the optimal policy $(\mu^*, \bar{\pi}^*)$ is composed of two greedy policies. The optimal stationary component, $\bar{\pi}^*$, is greedy with respect to the optimal exponential Q-function $Q^{\gamma}_*$, while the optimal initial policy, $\mu^*$, is greedy with respect to the optimal QH Q-function $Q^{\sigma,\gamma}_*$.

\subsubsection{QH Q-Learning Algorithm}

To learn these optimal Q-functions without a model, we propose QH Q-Learning. The algorithm maintains two concurrent estimates: (i)  $Z_n$ which learns $Q^{\gamma}_*$, and (ii)  $Q_n$, which learns the optimal QH Q-function $Q^{\sigma,\gamma}_*.$
Algorithm \ref{alg:qh_Q_learning} provides the full details of the update rules. The convergence of this procedure is guaranteed by the following theorem.

\begin{theorem}[Convergence of QH Q-Learning]
\label{th:qh_q_learning_convergence}
Suppose \ref{a:stepsize} and \ref{a:reward} are true. The iterates $(Z_n, Q_n)$ generated by the QH Q-learning algorithm (Algorithm \ref{alg:qh_Q_learning}) converge almost surely to the optimal action-value functions $(Q^{\gamma}_*, Q^{\sigma,\gamma}_*)$, respectively.
Furthermore, the policy $(\mu^*,\bar{\pi}^*)$ obtained by acting greedily with respect to these limit points is optimal:
\[
    \mu^*(s) = \arg\max_{a} Q^{\sigma,\gamma}_*(s,a)  \text{ and }  \pi^*(s) = \arg\max_{a} Q^{\gamma}_*(s,a).
\]
\end{theorem}

The proof follows a similar procedure to that of Theorem~\ref{th:policy_evaluation}.

\section{Experiments}
\label{sec:experiments}

\subsection{Inventory Control problem setup}
\label{subsec:InvCtrl_Setup}

In this subsection, we describe the inventory control problem which is used to run our simulations. We model the inventory control problem as an MDP, where the objective is to minimize the expected long-term cost. 

The inventory has a maximum capacity of $M$, and the state of the system at any given time represents the number of items available in stock on that day. Thus, the state space is given by $\mathcal{S} = \{0, 1, \dots, M\}$. The action space is $\mathcal{A} = \{0, 1, \dots, M\}$, where an action $a \in \mathcal{A}$ denotes the number of items procured on that day.  The immediate cost consists of three components: (i) the procurement cost incurred when purchasing items, (ii) the holding cost associated with storing unsold inventory, and (iii) the revenue generated from selling items. Let $c$ denote the unit cost of purchasing an item, $h$ represent the holding cost per unit of unsold inventory, and $p$ denote the unit selling price. Let $s_t, a_t,$ and $d_t$ denote the state, action, and demand on day $t$. We define $\hat{s}_t = \min(s_t + a_t,M)$, the total number of items available on day $t$ after procurement. Then, the next state is given by  
$s_{t+1} = \max(\hat{s}_t - d_t, 0),$
and the immediate reward received on day $t$ is given by $r(s_t, a_t, d_t) = -c \times a_t - h \times \max(\hat{s}_t - d_t, 0)  + p \times \min(\hat{s}_t, d_t).$

In this work, we consider an inventory system with parameters $M = 2, c = 5, h = 2$ and $p = 9$. The daily demand is a random variable with values 0, 1, 2 with probabilities 0.2, 0.3 and 0.5. We set $\sigma = 0.3$ and $\gamma = 0.9$.

\subsection{Optimal Policy}


In this experiment, we verify that our proposed QH Q-Learning algorithm correctly identifies the optimal policy. We first compute the ground-truth optimal policy using a model-based dynamic programming approach and then confirm that our QH Q-learning algorithm converges to the same solution.

\noindent\textbf{Ground-Truth Computation:}
Following the theoretical structure derived in Section~\ref{sec:optimal_control}, we first computed the optimal exponential value function, $V^*_{\gamma}$, using standard value iteration. An optimal stationary policy, $\pi^*$, is the greedy policy with respect to $V^*_{\gamma}$. Subsequently, an optimal initial policy, $\mu^*$, was determined. This procedure yielded optimal policy pair $(\mu^*, \bar{\pi}^*)$, where the initial policy $\mu^*$ selects actions $(1, 0, 0)$ for states $(0, 1, 2)$ respectively, and the stationary policy $\pi^*$ selects actions $(2, 1, 0)$ for the same states.



\noindent\textbf{Model-Free Validation:}We then ran the QH Q-Learning algorithm (Algorithm~\ref{alg:qh_Q_learning}) in the same environment without access to the model. The algorithm successfully recovered the exact same policy pair $(\mu^*, \bar{\pi}^*)$, validating the correctness of our model-free approach. Table~\ref{tab:opt_pol_Q_values} presents the converged action-value functions learned by the algorithm. Acting greedily with respect to the learned QH Q-function, $Q^{\sigma,\gamma}_*$, yields the optimal initial policy $\mu^*$, while acting greedily with respect to the learned exponential Q-function, $Q^{\gamma}_*$, yields the optimal stationary policy $\pi^*$.

\begin{table}[t]
\caption{Learned optimal action-value functions $Q^{\sigma,\gamma}_*$ and $Q^{\gamma}_*$.}
\centering
\renewcommand{\arraystretch}{1.2}
\begin{tabular}{ |c|ccc|ccc| } 
 \hline
 \multirow{2}{*}{\backslashbox{$s$}{$a$}} & \multicolumn{3}{c|}{$Q^{\sigma,\gamma}_*$} & \multicolumn{3}{c|}{$Q^{\gamma}_*$} \\
 \cline{2-7}
 & 0 & 1 & 2 & 0 & 1 & 2 \\
 \hline
 \hfil0 & \hfil 9.31 &  \hfil \textbf{11.38} &  \hfil 10.55 & \hfil 31.05 &  \hfil 33.75 &  \hfil \textbf{34.50} \\
 \hfil1  & \hfil\textbf{16.38} &  \hfil 15.55 & \hfil 10.55 & \hfil 38.75 &  \hfil \textbf{39.50} & \hfil 34.50 \\
 \hfil2 & \hfil\textbf{20.55} & \hfil 15.55 & \hfil 10.55 & \hfil\textbf{44.50} & \hfil 39.50 & \hfil 34.50 \\
 \hline
\end{tabular}
\label{tab:opt_pol_Q_values}
\end{table}

\subsection{Policy Evaluation}





In this subsection, we empirically validate the proposed policy evaluation method (Algorithm \ref{alg:policy_evaluation}). We test its convergence on the inventory control environment for a target policy class of interest: one-step non-stationary policies. To thoroughly test the algorithm's off-policy capabilities, we evaluate its performance in three distinct scenarios with mismatch between the target and behavior policies.

In all three configurations, the data is generated using a single behavior policy, $\nu = \psi$, where $\psi$ is the policy that selects actions uniformly at random in every state. The scenarios differ in their choice of the target policy, as follows:
\begin{enumerate}
    \item \textbf{Fully Off-Policy:} The target policy is the optimal policy, $(\mu^*, \bar{\pi}^*)$. In this case, both the initial and stationary components of the target policy differ from the uniform behavior policy.
    \item \textbf{Off-Policy Initial Step:} The target policy is $(\mu^*, \bar{\psi})$. Here, the stationary component ($\psi$) is learned on-policy, as it matches the behavior policy, while the initial component ($\mu^*$) is learned off-policy.
    \item \textbf{Off-Policy Stationary Steps:} The target policy is $(\psi, \bar{\pi}^*)$. This creates the opposite scenario, where the initial policy is on-policy and the stationary component ($\bar{\pi}^*$) is off-policy.
\end{enumerate}

\begin{figure}[t]
\centering
\includegraphics[width=\linewidth]{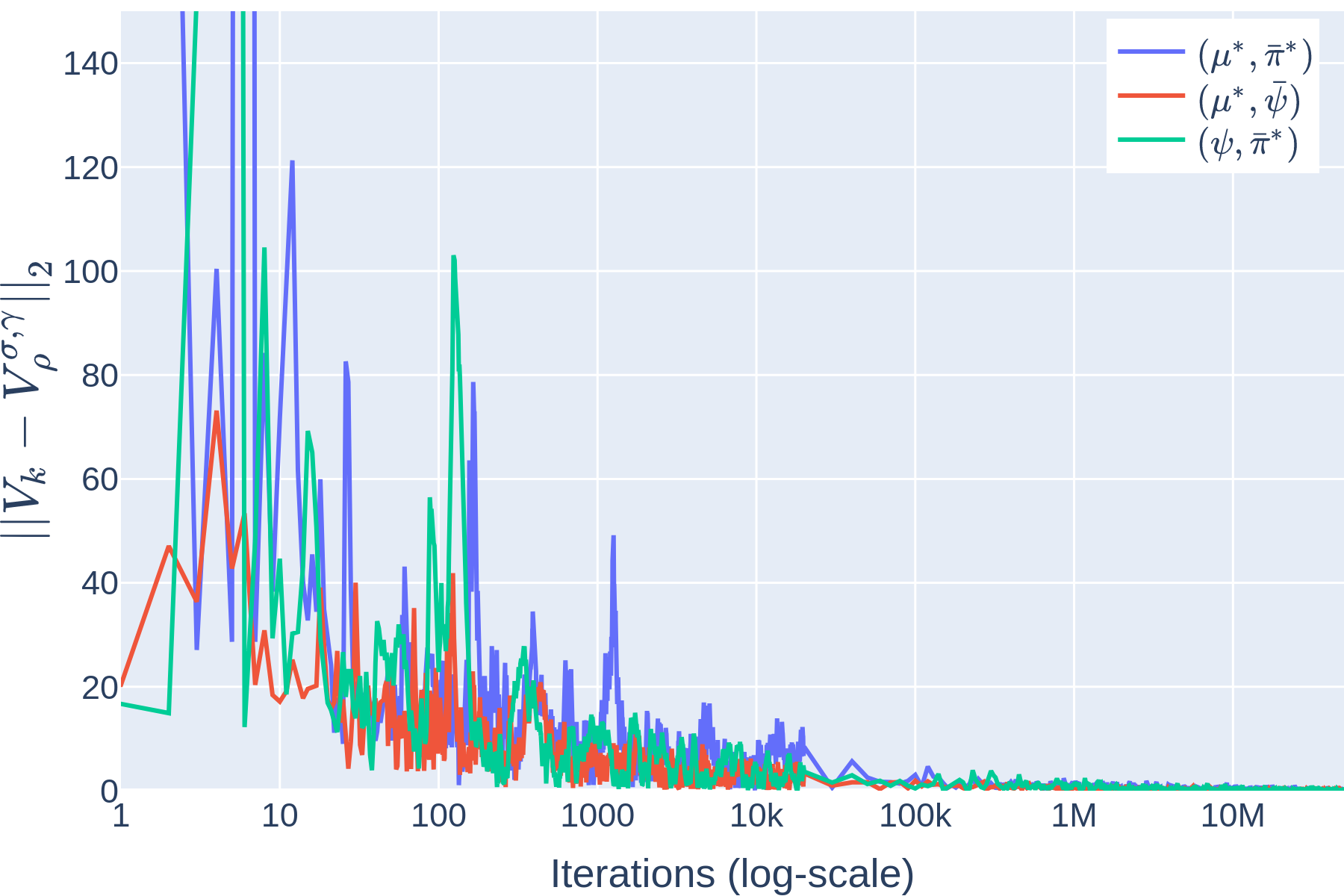}
\caption{The convergence of the Policy Evaluation algorithm (Algorithm \ref{alg:policy_evaluation}), showing the L2-norm error $\|V_k - V^{\sigma,\gamma}_{(\mu,\bar{\pi})}\|_2$ versus iterations on a log-scale.}
\label{fig:policy_eval_convergence}
\end{figure}

For each configuration, we ran Algorithm~\ref{alg:policy_evaluation} and tracked the L2-norm of the error between the estimate $V_k$ and the true value function $V^{\sigma,\gamma}_{(\mu,\bar{\pi})}$. Figure~\ref{fig:policy_eval_convergence} plots this error against the number of iterations on a logarithmic scale. As shown in the figure, the value function estimates converge to the true values in all three cases, providing strong empirical validation for our theoretical claims.

\section{Conclusion}
In this work, we present a comprehensive analysis of QH discounting for model-free reinforcement learning. Our foundational contribution is a structural result demonstrating that any policy's value is equivalent to that of a one-step non-stationary policy. This insight allowed us to characterize the optimal policy and develop two practical, provably convergent algorithms: one for policy evaluation and a novel QH Q-Learning algorithm for optimal control. 

By providing the first complete model-free solution for precommitted QH agents, this paper bridges a critical gap between behavioral economics and applied RL. The results presented here lay the groundwork for building agents that better align with human-like temporal preferences.

\section{Acknowledgements}
We wish to thank Prof. Gugan Thoppe (IISc, Bengaluru) for his invaluable guidance, and Anik Kumar Paul, T Nagesh, and Ananyabrata Barua of the Vivitsu Lab at IISc, Bengaluru for their helpful discussions and feedback.

\bibliography{references}
\bibliographystyle{abbrv}

\addtolength{\textheight}{-12cm}   

\end{document}